%% file: cosp_2fwl_gnns.tex
\documentclass[letterpaper]{article} 
\usepackage{aaai2026}  
\usepackage{times}  
\usepackage{helvet}  
\usepackage{courier}  
\usepackage[hyphens]{url}  
\usepackage{graphicx} 
\usepackage{natbib}  
\usepackage{caption} 
\usepackage{algorithm}
\usepackage{algorithmic}

\usepackage{newfloat}
\usepackage{listings}
\DeclareCaptionStyle{ruled}{labelfont=normalfont,labelsep=colon,strut=off} 
\floatstyle{ruled}
\newfloat{listing}{tb}{lst}{}
\floatname{listing}{Listing}
\input{math_commands.tex}

\usepackage{multirow}
\usepackage{enumitem}
\usepackage{booktabs}
\usepackage{amssymb}
\usepackage{amsmath}
\usepackage{amsthm}

\title{Connectivity-Guided Sparsification of 2-FWL GNNs: Preserving Full Expressivity with Improved Efficiency}

\author{
    Rongqin Chen\textsuperscript{\rm 1},
    Fan Mo\textsuperscript{\rm 2, \rm 3},
    Pak Lon Ip\,\textsuperscript{\rm 1}, 
    Shenghui Zhang\textsuperscript{\rm 1}, 
    Dan Wu\textsuperscript{\rm 4},
    Ye Li\textsuperscript{\rm 4, \rm 5}, 
    Leong Hou U\textsuperscript{\rm 1}\thanks{Corresponding author.}
}
\affiliations{
    \textsuperscript{\rm 1}University of Macau\\
    \textsuperscript{\rm 2}ZOZO Research\\
    \textsuperscript{\rm 3}Waseda University\\
    \textsuperscript{\rm 4}Shenzhen Institutes of Advanced Technology, Chinese Academy of Sciences\\
    \textsuperscript{\rm 5}Faculty of Computer Science and Control Engineering, Shenzhen University of Advanced Technology\\

    \{chen.rongqin, zhang.shenghui\}@connect.um.edu.mo; bakubonn@toki.waseda.jp \{dan.wu ye.li\}@siat.ac.cn \{paklonip, ryanlhu\}@um.edu.mo
}

\usepackage{bibentry}

\begin{document}

\maketitle

\begin{abstract}
    Higher-order Graph Neural Networks (HOGNNs) based on the 2-FWL test achieve superior expressivity by modeling 2- and 3-node interactions, but at $\mathcal{O}(n^3)$ computational cost. However, this computational burden is typically mitigated by existing efficiency methods at the cost of reduced expressivity. We propose \textbf{Co-Sparsify}, a connectivity-aware sparsification framework that eliminates \emph{provably redundant} computations while preserving full 2-FWL expressive power. Our key insight is that 3-node interactions are expressively necessary only within \emph{biconnected components}—maximal subgraphs where every pair of nodes lies on a cycle. Outside these components, structural relationships can be fully captured via 2-node message passing or global readout, rendering higher-order modeling unnecessary. Co-Sparsify restricts 2-node message passing to connected components and 3-node interactions to biconnected ones, removing computation without approximation or sampling. We prove that Co-Sparsified GNNs are as expressive as the 2-FWL test. Empirically, on PPGN, Co-Sparsify matches or exceeds accuracy on synthetic substructure counting tasks and achieves state-of-the-art performance on real-world benchmarks (ZINC, QM9). This study demonstrates that high expressivity and scalability are not mutually exclusive: principled, topology-guided sparsification enables powerful, efficient GNNs with theoretical guarantees.
\end{abstract}

\begin{links}
    \link{Code}{https://github.com/RongqinChen/HOGNN-Sparsify}
\end{links}

\section{Introduction}

Graph Neural Networks (GNNs) serve as the predominant framework for learning on graph-structured data.
A central challenge is enhancing their \emph{expressive power}: the ability to distinguish non-isomorphic graphs and detect fine-grained structural patterns.
Standard message-passing GNNs—such as GCN~\cite{kipf2017semisupervised}, GIN~\cite{xu2019powerful}, and GAT~\cite{velickovic2018graph}—are limited by the expressivity of the 1-WL test. Higher-order GNNs (HOGNNs), inspired by the $k$-dimensional Weisfeiler-Leman ($k$-WL) and $k$-Folklore WL ($k$-FWL) hierarchies~\cite{morris2019weisfeiler,maron2019provably}, overcome this by performing message passing over $k$-tuples of nodes. These models strictly subsume classical GNNs—and even recent architectures like Graph Transformers~\cite{rampásek2022recipe}—in expressive power.

Yet, HOGNNs suffer from combinatorial complexity. Time and memory scale with the order $k$, making higher orders impractical.
Among them, 2-FWL GNNs strike a practical balance: they match the expressivity of the 3-WL test while remaining implementable~\cite{maron2019provably}.
Still, they require $\mathcal{O}(n^3)$ memory for an $n$-node graph.
Even optimized variants like PPGN~\cite{maron2019provably}, which use batched matrix operations, incur $\mathcal{O}(\eta^2)$ per-graph memory due to padding, where $\eta$ is the largest graph size in a batch.
This limits their use on large or irregular graphs.

\begin{figure}[!h]
    \centering
    \includegraphics[width=1.0\linewidth]{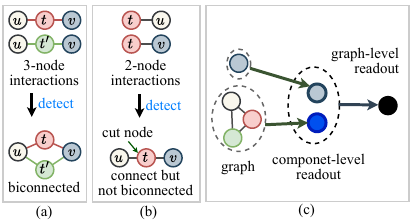}
    \caption{\textbf{Principle of connectivity-aware sparsification.} (a) Biconnected case: 3-node interactions needed to distinguish paths. (b) Cut-node case: 2-node interactions suffice; 3-node redundant. (c) Disconnected case: component-level structural properties (e.g., component size) are captured by component-level readout and global structural properties (e.g., component count) are captured by graph-level readout, making explicit 2- or 3-node modeling unnecessary.}
    \label{fig:illustration}
\end{figure}

Prior efficiency methods—subgraph sampling (e.g., ESAN~\cite{zhao2022from}), set-based reduction (e.g., KCSetGNN~\cite{zhao2022practical}), or localized aggregation (e.g., 1-2-3-GNN~\cite{morris2019weisfeiler})—trade expressivity for efficiency. They rely on approximation, sampling, or restricted neighborhoods. We ask instead:
\textbf{Can we improve efficiency by eliminating only computations that are redundant for expressivity—without any loss in expressive power?}

We focus on 2-FWL GNNs, which update each node pair $(u,v)$ by aggregating messages from coupled pairs $((u,t), (t,v))$ over all $t \in V$.
When $u$, $t$, and $v$ are distinct, this forms a \emph{3-node interaction}—critical for capturing structures like two internally disjoint paths\footnote{$u$-$v$ paths are \emph{internally disjoint} if they share no intermediate nodes—e.g., $u \to v$, $u \to t_1 \to v$, and $u \to t_2 \to v$.}, enabling cycle and biconnectivity detection.
When $u = t$ or $t = v$, this interaction reduces to a \emph{2-node interaction}, capturing only whether paths exist between $u$ and $v$.

Our key insight is that \textbf{3-node interactions are expressively necessary only within biconnected components}. By Menger’s theorem~\cite{goring2000short}, any two nodes in the same biconnected component are connected by at least two internally disjoint paths (Figure~\ref{fig:illustration}(a)). Only in this case does modeling triplets \((u,t,v)\) provide additional expressive power.
Outside biconnected components, 3-node interactions are redundant. First, all \(u\)-\(v\) paths must pass through a \emph{cut node} \(t\), decomposing the connection into \(u\)-\(t\) and \(t\)-\(v\) subpaths. Since 2-FWL already captures all pairwise interactions, aggregating over \((u,t,v)\) yields no expressivity gain (Figure~\ref{fig:illustration}(b)). Second, for \emph{disconnected} node pairs, structural context—such as component count or size—is already encoded by standard readout functions, rendering explicit 2- or 3-node modeling unnecessary (Figure~\ref{fig:illustration}(c)).

We propose \textbf{Co-Sparsify}, a connectivity-aware sparsification scheme that preserves the full expressivity of the 2-FWL test by eliminating only provably redundant computations. The method restricts 2-node interactions to node pairs within the same connected component and 3-node interactions to triples within the same biconnected component. These structural constraints ensure that all removed operations contribute no additional expressive power.
No sampling, approximation, or heuristic pruning is applied. Instead, sparsification is guided by exact graph topology.
The preprocessing overhead is negligible. Connected components are identified via breadth-first search, and biconnected components via block-cut tree decomposition~\cite{hopcroft1973algorithm}. For a graph with $n$ nodes and $m$ edges, the entire step runs in $O(n + m)$ time.

We prove that Co-Sparsified GNNs are as expressive as the 2-FWL test under standard assumptions—injective aggregation and consistent initialization. The framework applies to any 2-FWL-based architecture, including PPGN~\cite{maron2019provably} and TGT~\cite{hussain2024triplet}. When applied to PPGN, Co-Sparsify correctly counts paths and cycles, confirming that expressivity is preserved. It also matches or exceeds the original model’s performance on both synthetic tasks and real-world benchmarks such as ZINC and QM9.

Our contributions include:
\begin{itemize}
    \item We identify that \emph{3-node interactions} in 2-FWL GNNs are expressively useful \emph{only within biconnected components}, and \emph{2-node interactions} matter \emph{only within connected components}.
    \item We propose a \emph{connectivity-aware message-passing scheme}---\textbf{Co-Sparsify}---that restricts 3-node interactions to biconnected components and 2-node interactions to connected components. Our method eliminates provably redundant computations without approximation, sampling, or heuristic pruning.
    \item We prove that Co-Sparsified GNNs are \emph{as powerful as the 2-FWL test} in distinguishing non-isomorphic graphs---making it the first sparsification method for HOGNNs with \emph{guaranteed expressivity preservation}.
    \item We demonstrate that Co-Sparsify achieves preserved or improved accuracy across synthetic substructure counting tasks and real-world graph benchmarks.
\end{itemize}

Our work shows that high expressivity need not require uniform, dense computation over all node tuples. Instead, by aligning message passing with \emph{structural criticality}---here, connected and biconnected components---we achieve efficiency through \emph{insight}, not approximation. This opens a principled path toward scalable, theoretically grounded GNNs that preserve expressive power by design.

\section{Preliminaries}
\label{sec:preliminaries}

This section introduces notation and background concepts used throughout the paper.

\noindent \textbf{Notation.}
We denote an undirected graph as $G = (V, E)$, where $V$ is a set of $n$ nodes and $E \subseteq V \times V$ is a set of $m$ edges. The adjacency matrix $A \in \{0,1\}^{n \times n}$ satisfies $A[u,v] = 1$ if $(u,v) \in E$, and $0$ otherwise. The degree matrix $D$ is diagonal, with $D[v,v] = \sum_{u} A[v,u]$.
The random walk probability matrix is computed as  $\widehat{A} = D^{-1}A$.
We represent multisets using $\lbbr \cdot \rbbr$.

\noindent \textbf{Graph Connectivity.}
A \emph{connected component} is a maximal subgraph in which all node pairs are mutually reachable. A \emph{biconnected component} is a maximal subgraph (on at least three nodes) that remains connected after removing any single node. \emph{Cut nodes}—whose removal increases the number of connected components—are shared across biconnected components and bridge edges. The \emph{block-cut tree} decomposition identifies these components efficiently in $O(n + m)$ time~\cite{hopcroft1973algorithm}. This hierarchical structure underpins our sparsification strategy.

\noindent \textbf{Expressive Power.}
Two graphs $G$ and $H$ are \emph{isomorphic}, denoted $G \simeq H$, if there exists a bijection $\pi: V_G \to V_H$ such that $(u,v) \in E_G$ if and only if $(\pi(u), \pi(v)) \in E_H$.
A GNN \emph{distinguishes} two non-isomorphic graphs if it produces different graph-level representations for them.
It \emph{captures} a substructure $Q$ if its representations reliably differ between graphs that contain $Q$ and those that do not.
The \emph{expressive power} of a GNN is defined by its ability to distinguish non-isomorphic graphs and detect such substructures.

\noindent \textbf{2-FWL GNNs.}
These models update node pair representations following the 2-FWL test. Let $\boldsymbol{h}^{(l)}(u,v)$ be the representation of pair $(u,v)$ at layer $l$. Initial features combine node, edge, and structural information:
\begin{equation}
    \boldsymbol{h}^{(0)}(u,v) = \left( \boldsymbol{x}(u), \boldsymbol{x}(v), \boldsymbol{e}(u,v), \boldsymbol{p}(u,v) \right),
\end{equation}
where $\boldsymbol{x}(\cdot)$ are node features, $\boldsymbol{e}(u,v)$ is the edge feature (zero if no edge), and $\boldsymbol{p}(u,v)$ is a \emph{structural encoding} (SE). A common SE is relative random walk probability (RRWP):
\[
    \boldsymbol{P}[u,v,:] = \left[ I, \widehat{A}, \widehat{A}^2, \dots \right][u,v,:].
\]
It encodes self-pair status, direct connectivity, and multi-hop reachability.

At layer $l$, each pair $(u,v)$ aggregates messages over intermediate nodes $t \in V$:
\begin{equation}
    \begin{aligned}
        \label{eq:standard_2fwl}
        \boldsymbol{t}^{(l)}(u, v)
         & = \Blbbr \phi^{(l)}\left(\boldsymbol{h}^{(l-1)}(u, t), \boldsymbol{h}^{(l-1)}(t, v) \right) \mid t \in V \Brbbr, \\
        \boldsymbol{h}^{(l)}(u,v)
         & = \Phi^{(l)} \left( \boldsymbol{h}^{(l-1)}(u,v),\, \text{AGG}\left(\boldsymbol{t}^{(l)}(u,v)\right) \right),
    \end{aligned}
\end{equation}
where $\Phi^{(l)}$ is learnable and $\text{AGG}$ is injective (e.g., sum or MLP-based). Full aggregation over all $t$ leads to $\mathcal{O}(n^3)$ memory and computation per layer.

To improve efficiency, PPGN~\cite{maron2019provably} uses batched matrix operations. Let $\boldsymbol{H}^{(l)} \in \mathbb{R}^{b \times d \times \eta \times \eta}$ be the batched pair representations, where $b$ is batch size, $d$ feature dimension, and $\eta$ the maximum graph size after padding. PPGN updates via:
\begin{equation}
    \boldsymbol{H}^{(l)} = \Phi^{(l)} \left( \boldsymbol{H}^{(l-1)},\, \boldsymbol{H}^{(l-1)} \circledast \boldsymbol{H}^{(l-1)} \right),
\end{equation}
where $\circledast$ denotes matrix multiplication along node dimensions. For graph $i$, this computes:
\begin{equation}
    \begin{aligned}
         & \boldsymbol{H}^{(l)}[i,:,u,v] =                                                                                                              \\
         & \Phi^{(l)} \left(\boldsymbol{H}^{(l-1)}[i,:,u,v], \sum_{t \in V} \boldsymbol{H}^{(l)}[i,:,u,t] \otimes \boldsymbol{H}^{(l)}[i,:,t,v] \right)
    \end{aligned}
\end{equation}
where $\otimes$ denotes element-wise multiplication.

Despite avoiding explicit loops, PPGN still requires $\mathcal{O}(\eta^2)$ memory per graph due to padding—limiting scalability on large or irregular graphs.

\noindent \textbf{Problem Setup.}
We aim to design sparsified 2-FWL GNNs that remove only computations provably redundant for expressivity. Our goal is to preserve full 2-FWL expressive power while improving efficiency. We prove that the resulting model matches 2-FWL expressivity and evaluate its performance on substructure counting and graph-level prediction tasks.

\section{Related Work}
\label{sec:related_work}

\paragraph{Expressive GNNs and the WL hierarchy.}
The expressive power of GNNs is often bounded by the 1-WL test, limiting their ability to distinguish non-isomorphic graphs~\cite{xu2019powerful}.
Standard models like GCN~\cite{kipf2017semisupervised}, GAT~\cite{velickovic2018graph}, and Gated GCN~\cite{li2016gated} fall into this class.
To surpass these limits, HOGNNs based on the $k$-WL and $k$-Folklore WL ($k$-FWL) hierarchies operate on $k$-tuples of nodes, achieving greater expressivity at $\mathcal{O}(n^k)$ cost~\cite{morris2019weisfeiler,maron2019provably}. Recent works align GNNs with 2-FWL via low-rank attention~\cite{puny2020graph}, often introducing architectural complexity to balance expressivity and efficiency.

\paragraph{Efficient HOGNNs.}
To reduce the cost of HOmodels, several directions have emerged.
Set-based approaches reformulate $k$-tuple learning as $k$-set learning, reducing the number of representations by exploiting permutation invariance~\cite{zhao2022practical}, while local message-passing methods restrict aggregation to neighborhoods~\cite{morris2020weisfeiler}. Subgraph-based methods enhance expressivity by lifting graphs into higher-dimensional spaces~\cite{bodnar2021weisfeiler,bouritsas2023improving}, though often without addressing scalability. Simplified spectral models achieve linear complexity~\cite{wu2019simplifying}, but may sacrifice structural fidelity through approximation.

\paragraph{Graph Sparsification.}
Sparsification reduces computational load by pruning non-essential edges or computations. Learning-based methods train edge masks to improve generalization~\cite{zheng2020robust,rathee2021learnt} or use optimization frameworks like ADMM~\cite{li2022graph}.
Others sparsify model weights via pruning or sparse training~\cite{peng2022towards}. While effective, these approaches require additional training and offer no guarantees on expressivity preservation. In contrast, our method, \emph{Co-Sparsify}, uses the graph’s connectivity structure—connected and biconnected components—to deterministically remove only expressivity-redundant interactions, with provable 2-FWL equivalence.

\paragraph{Hierarchical GNNs.}
Hierarchical models capture multi-scale structure via pooling or auxiliary layers.
DiffPool learns soft cluster assignments for hierarchical representation~\cite{ying2018hierarchical}, while HGNet organizes graphs into layers for logarithmic message passing~\cite{rampavsek2021hierarchical}. These methods enhance expressivity but add architectural complexity. Our approach differs by simplifying computation—leveraging structural hierarchies (e.g., block-cut trees) to sparsify message passing without added parameters or layers.

\paragraph{Our Position in the Literature.}
While prior work has made significant progress in improving the efficiency of HOGNNs, most approaches either reduce expressivity or still incur high computational costs.
In contrast, we propose a novel sparsification strategy for 2-FWL GNNs that preserves full expressive power while reducing runtime. Our method leverages insights into graph connectivity to identify and retain only the most structurally informative interactions, achieving both theoretical guarantees and strong empirical performance.

\section{Method}
\label{sec:method}

We present \textbf{Co-Sparsify}, a connectivity-aware sparsification framework for HOGNNs. It eliminates only expressivity-redundant computations while preserving full 2-FWL expressive power, yielding \textbf{Co-Sparsified 2-FWL GNNs}.
Leveraging connected component and block-cut tree decompositions, Co-Sparsify restricts 3-node interactions to biconnected components and 2-node modeling to connected components—reducing computation without compromising substructure detection.

We provide proof sketches for our theoretical results and defer the detailed proofs to Appendix~\ref{section:proofs_of_lemmas_and_theorems}.

\begin{figure}[!h]
    \centering
    \includegraphics[width=1\columnwidth]{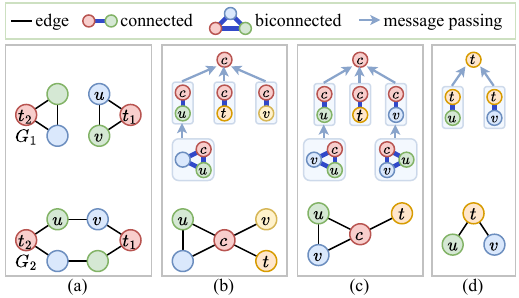}
    \caption{\textbf{Justification for connectivity-aware sparsification.} (a) Distinguishing graphs with different path multiplicities (e.g., $u \to v$ vs. $u \to t \to v$) requires 3-node interactions. (b) Cut node $c$ separates $u$, $t$, and $v$ into distinct components; pairwise interactions $(u,c)$, $(t,c)$, $(v,c)$ suffice to determine the structure. (c) Cut node $c$ separates $t$ from $\{u,v\}$, with $u$ and $v$ connected; structure is captured by $(u,c)$, $(t,c)$, and $(v,c)$. (d) Cut node $t$ separates $u$ and $v$; interactions $(u,t)$ and $(v,t)$ fully determine the graph.}
    \label{fig:justification}
\end{figure}

\subsection{Connectivity-Guided Sparsification Principle}

The expressive power of 2-FWL GNNs stems from modeling 2-node and 3-node interactions. However, we show that such interactions are only expressively necessary within connected and biconnected components, respectively. Outside these regions, higher-order interactions contribute no additional discriminative power and can be provably removed without loss of expressivity. This principle is formalized in lemmas.

\begin{lemma}[Necessity of 3-Node Interactions in Biconnected Components]
    \label{lem:biconnected_necessity}
    Let $u$ and $v$ be distinct nodes in a graph $G$. If $u$ and $v$ lie in the same biconnected component, then there are at least two internally disjoint paths between them. Detecting such configurations requires 3-node interactions and cannot be captured by 2-node interactions alone.
\end{lemma}

\begin{proof}
    By Menger’s theorem, biconnectedness implies multiple internally disjoint paths. A 3-node interaction $(u,t,v)$ indicates whether $t$ lies on a path from $u$ to $v$, and aggregating over $t$ allows the model to detect path multiplicity---essential for detecting biconnectedness.
\end{proof}

As shown in Figure~\ref{fig:justification}(a), the node pairs $(u, v)$ in $G_1$ and $G_2$ are connected by multiple internally disjoint paths. While 1-WL cannot distinguish these graphs or pairs, 2-FWL can, illustrating the necessity of 3-node interactions.

\begin{lemma}[Redundancy of 3-Node Interactions Across Cut Nodes]
    \label{lem:cut_redundancy}
    Let $u$, $t$, and $v$ be distinct nodes in a connected graph $G$. If every path from $u$ to $v$ through $t$ passes through a cut node separating $t$ from $\{u,v\}$, then the 3-node interaction $(u,t,v)$ is expressivity-redundant: its contribution to $\boldsymbol{h}(u,v)$ is fully captured by 2-node interactions $(u,t)$ and $(t,v)$, and removing it preserves 2-FWL expressivity.
\end{lemma}

\begin{proof}
    Let $c$ be a cut node separating $t$ from $\{u,v\}$. After removing $c$, either:
    (i) $u$, $t$, $v$ are in three components: then $\boldsymbol{h}(u,c)$, $\boldsymbol{h}(t,c)$, $\boldsymbol{h}(v,c)$ encode all structural roles; or
    (ii) $u$ and $v$ remain connected: then $u,v,c$ lie in one biconnected component,
    and 3-node interactions within it (e.g., $(u,c,v)$) capture local structure, while $(t,c)$ is updated separately.
    If $t$ is the cut node, $\boldsymbol{h}(u,t)$ and $\boldsymbol{h}(t,v)$ are computed independently across blocks.
    In all cases, the interaction $(u,t,v)$ only combines precomputed, separable pairwise information and contributes no new expressive power.
\end{proof}

\begin{lemma}[Irrelevance of Interactions Between Disconnected Components]
    \label{lem:disconnected_irrelevance}
    Let $u$ and $v$ be nodes in different connected components. Then no path exists between them, and their structural independence is implicitly encoded. Explicit modeling of 2-node or 3-node interactions involving both components is unnecessary for substructure detection, provided that component-level readouts capture macro-topological features.
\end{lemma}

\begin{proof}
    Disconnected pairs have no shared paths. The absence of interaction itself encodes disconnection. Component-level readouts aggregate over intra-component pairs, capturing local structure (e.g., cycles), while graph-level aggregation detects global properties (e.g., number of triangles). No inter-component interaction is needed to detect any connected subgraph.
\end{proof}

\subsection{Co-Sparsified Neighborhood Construction}
\label{sec:neighborhood}

To implement our sparsification principle, we define a \emph{co-sparsified neighbor set} $\mathcal{N}_{\text{sp}}(u,v)$ for each node pair $(u,v)$, based on the graph’s connectivity structure. We first decompose $G$ into connected components $\mathcal{C}$ and biconnected components (blocks of at least three nodes) $\mathcal{B}$ using Tarjan’s algorithm~\cite{hopcroft1973algorithm}, which runs in $O(n + m)$ time.

If $u$ and $v$ are in different connected components, we set $\mathcal{N}_{\text{sp}}(u,v) = \emptyset$, as they are structurally independent—this disconnection is captured at the global level through readout. Otherwise, for $u, v$ in the same component $C \in \mathcal{C}$, $\mathcal{N}_{\text{sp}}(u,v)$ includes only interactions that are expressively necessary.

We include 3-node interactions $\big((u,t), (t,v)\big)$ only when $u$, $t$, and $v$ are distinct and lie within the same biconnected block $B \in \mathcal{B}$, ensuring aggregation occurs precisely where path multiplicity and cyclic structure must be resolved (Lemma~\ref{lem:biconnected_necessity}). For 2-node interactions, we include $\big((u,u), (u,v)\big)$ and $\big((u,v), (v,v)\big)$ to propagate information from self-pairs (i.e., nodes) to pairs. Additionally, for $u \neq v$, we include $\big((v,u), (u,v)\big)$ in $\mathcal{N}_{\text{sp}}(v,v)$ to propagate messages from pairs back to nodes. Finally, for self-pairs $(u,u)$, we include $\big((u,u), (u,u)\big)$ to ensure non-empty neighborhoods and stable updates.

This construction preserves all interactions needed for full 2-FWL expressivity, while eliminating only those proven redundant.

\subsection{Co-Sparsified Message Passing Scheme}
\label{sec:message_passing}

Using $\mathcal{N}_{\text{sp}}(u,v)$, we define the sparse 2-FWL message passing update. Initial representations follow Section~\ref{sec:preliminaries}:
\[
    \boldsymbol{h}^{(0)}_{\text{sp}}(u,v) = \left( \boldsymbol{x}(u), \boldsymbol{x}(v), \boldsymbol{e}(u,v), \boldsymbol{p}(u,v) \right).
\]
At layer $l$, for $(u,v)$ in the same connected component:
\begin{equation}
    \label{equation:co_sparsified}
    \begin{aligned}
        \boldsymbol{t}^{(l)}_{\text{sp}}(u, v)
         & = \Blbbr \phi^{(l)} \left( \boldsymbol{h}^{(l-1)}_{\text{sp}}(u, t), \boldsymbol{h}^{(l-1)}_{\text{sp}}(t, v) \right)                \\
         & \qquad \mid \big((u,t), (t,v)\big) \in \mathcal{N}_{\text{sp}}(u,v) \Brbbr,                                                          \\
        \boldsymbol{h}^{(l)}_{\text{sp}}(u,v)
         & = \Phi^{(l)} \left( \boldsymbol{h}^{(l-1)}_{\text{sp}}(u,v),\, \text{AGG}\left(\boldsymbol{t}^{(l)}_{\text{sp}}(u,v)\right) \right),
    \end{aligned}
\end{equation}
where $\text{AGG}$ is an injective aggregation (e.g., sum, mean, or MLP-based), and $\Phi^{(l)}$ is a learnable function (e.g., MLP). For disconnected pairs, no update is performed.

\subsection{Node-Level and Graph-Level Readouts}

For node-level tasks, the representation $\boldsymbol{z}_v$ is computed via readout over incoming pair representations within $v$'s connected component $C$:
\[
    \boldsymbol{z}_v = \varPsi\left(\text{AGG}\left(\Blbbr \boldsymbol{h}^{(L)}_{\text{sp}}(u,v) \mid u \in C \Brbbr \right)\right).
\]
For graph-level tasks, we first compute component-level representations. For each connected component $C$, we apply two parallel readouts over self-pairs and off-diagonal pairs:
\begin{equation}
    \begin{aligned}
        \boldsymbol{z}_{C} = \varPhi & \left( \text{AGG}\left(\Blbbr \boldsymbol{h}^{(L)}_{\text{sp}}(u,u) \mid u \in C \Brbbr\right), \right.                 \\
                                     & \;\; \left. \text{AGG}\left(\Blbbr\boldsymbol{h}^{(L)}_{\text{sp}}(u,v) \mid u,v \in C, u \neq v \Brbbr \right)\right).
    \end{aligned}
\end{equation}
The final graph representation is:
\[
    \boldsymbol{z}_G = \Psi \left( \text{AGG}\left(\Blbbr \boldsymbol{z}_C \mid C \in G \Brbbr \right) \right).
\]
If $G$ is connected, $\boldsymbol{z}_G = \boldsymbol{z}_C$ for the single component.

\subsection{Expressive Power Study}
\label{sec:expressiveness}

We prove that \textbf{Co-Sparsified 2-FWL GNNs are as powerful as the 2-FWL test} in detecting substructures and distinguishing non-isomorphic graphs. Since graph isomorphism is a special case of substructure detection, this establishes full 2-FWL expressive power.

\begin{theorem}
    \label{thm:subgraph_expressiveness}
    Let $Q$ be a query subgraph. A Co-Sparsified 2-FWL GNN can detect $Q$ if and only if a standard 2-FWL GNN can, under injective aggregation and consistent initialization. Thus, the two models are \emph{equally expressive} in substructure detection.
\end{theorem}

\begin{proof}
    The expressive power of 2-FWL stems from its ability to model 2-node and 3-node interactions, enabling it to count paths, cycles, and other symmetric patterns. Co-Sparsify uses the same update rules but restricts 3-node interactions to biconnected components and 2-node interactions to connected components.

    We show this sparsification preserves all essential structural information. When a subgraph $Q$—like a triangle or 4-cycle—lies within a biconnected component, all required 3-node interactions are retained, so $Q$ is detected exactly as in 2-FWL. When $Q$ spans multiple blocks (e.g., a path through a cut node), its structure decomposes into segments $(u,t)$ and $(t,v)$. Since 2-FWL already updates all such pairs, their representations $\boldsymbol{h}(u,t)$ and $\boldsymbol{h}(t,v)$ fully encode the path—aggregating over $(u,t,v)$ adds no new information and can be safely omitted.

    For disconnected queries, global topology (e.g., component count, size) is captured via unchanged readout operations. Crucially, the ability to distinguish complex patterns—such as two internally disjoint paths—depends on biconnected structure, which our method fully preserves.

    By induction on message-passing layers, and under injective aggregation, Co-Sparsify computes pair representations that induce the same distinctions as 2-FWL. It detects the same subgraphs and separates the same graph classes—proving that only redundant computations are removed.
\end{proof}

\begin{figure}[!h]
    \centering
    \includegraphics[width=1.0\columnwidth]{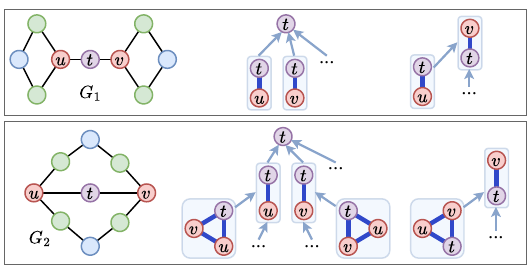}
    \caption{\textbf{Co-Sparsified 2-FWL's expressive power.} It distinguishes graphs $G_1$ and $G_2$—indistinguishable by 1-WL but distinguishable by 2-FWL—by capturing structural differences in node and pair representations, enabling correct subgraph detection.\label{fig:expressivity}}
\end{figure}

As shown in Figure~\ref{fig:expressivity}, Co-Sparsified 2-FWL GNNs achieve the same expressive power as standard 2-FWL GNNs on $G_1$ and $G_2$.

\subsection{Computational Efficiency}

Standard 2-FWL GNNs update $O(n^2)$ 2-tuples using $O(n^3)$ 3-tuple interactions. Our method reduces this to $O\left(\sum_{i} n_i^2\right)$ pairs (summing over connected components of size $n_i$) and $O\left(\sum_{j} n_{b_j}^3\right)$ triples (summing over biconnected components of size $n_{b_j}$). In graphs with sparse or fragmented higher-order connectivity (e.g., molecular), this results in significant reductions---from cubic to sub-quadratic or sub-cubic complexity---without expressivity loss.

\section{Experiments}
\label{sec:experiments}

\begin{table*}[!h]
    \centering
    \renewcommand{\familydefault}{\rmdefault}
    \fontsize{9pt}{9pt}\selectfont
    \setlength{\tabcolsep}{1.5pt}
    \begin{tabular}{c|cccccccc}
        \toprule
        Dataset      & Counting & QM9     & ZINC-12k & ZINC-250k & FRANK. & NCI1  & NCI109 & ENZYMES \\
        \midrule
        \#Graphs     & 5,000    & 130,831 & 12,000   & 249,456   & 4,337  & 4,110 & 4,127  & 600     \\
        Avg. \#Nodes & 18.8     & 18.0    & 23.2     & 23.1      & 16.9   & 29.9  & 29.7   & 32.6    \\
        Avg. \#Edges & 31.3     & 18.7    & 24.9     & 24.9      & 17.9   & 32.3  & 32.1   & 62.1    \\
        \bottomrule
    \end{tabular}
    \caption{Dataset statistics.\label{table:dataset}}
\end{table*}

\begin{table*}[!h]
    \centering
    \renewcommand{\familydefault}{\rmdefault}
    \fontsize{9pt}{9pt}\selectfont
    \setlength{\tabcolsep}{1.2mm}
    \begin{tabular}{c|rrrrrrrr|r}
        \toprule
        Target                      & DTNN       & MPNN    & PPGN       & NGNN    & KP-GIN$'$ & I$^{2}$-GNN & N$^{2}$-GNN  & PPGN+RRWP    & CoSp-PPGN+RRWP \\ \midrule
        $\mu$                       & \bf{0.244} & 0.358   & \bf{0.231} & 0.433   & 0.358     & 0.428       & 0.333        & 0.332        & 0.321          \\
        $\alpha$                    & 0.95       & 0.89    & 0.382      & 0.265   & 0.233     & 0.230       & \bf{0.193}   & 0.200        & \bf{0.183}     \\
        $\varepsilon_{\text{HOMO}}$ & 0.00388    & 0.00541 & 0.00276    & 0.00279 & 0.00240   & 0.00261     & \bf{0.00217} & 0.00236      & \bf{0.00214}   \\
        $\varepsilon_{\text{LUMO}}$ & 0.00512    & 0.00623 & 0.00287    & 0.00276 & 0.00236   & 0.00267     & \bf{0.00210} & 0.00231      & \bf{0.00210}   \\
        $\Delta \varepsilon$        & 0.0112     & 0.0066  & 0.00406    & 0.00390 & 0.00333   & 0.00380     & \bf{0.00304} & 0.00327      & \bf{0.00299}   \\
        $\langle R^{2} \rangle$     & 17.0       & 28.5    & 16.7       & 20.1    & 16.51     & 18.64       & \bf{14.47}   & 16.24        & \bf{14.25}     \\
        ZPVE                        & 0.00172    & 0.00216 & 0.00064    & 0.00015 & 0.00017   & 0.00014     & 0.00013      & \bf{0.00012} & \bf{0.00012}   \\
        $U_{0}$                     & 2.43       & 2.05    & 0.234      & 0.205   & 0.0682    & 0.211       & 0.0247       & \bf{0.0111}  & \bf{0.0149}    \\
        $U$                         & 2.43       & 2.00    & 0.234      & 0.200   & 0.0696    & 0.206       & 0.0315       & \bf{0.01175} & \bf{0.01373}   \\
        $H$                         & 2.43       & 2.02    & 0.229      & 0.249   & 0.0641    & 0.269       & 0.0182       & \bf{0.01266} & \bf{0.01637}   \\
        $G$                         & 2.43       & 2.02    & 0.238      & 0.253   & 0.0484    & 0.261       & 0.0178       & \bf{0.01278} & \bf{0.01767}   \\
        $C_{v}$                     & 0.27       & 0.42    & 0.184      & 0.0811  & 0.0869    & \bf{0.0730} & \bf{0.0760}  & 0.08589      & 0.07619        \\
        \bottomrule
    \end{tabular}
    \caption{MAE results on QM9 (top 2 results are bold).\label{table:qm9}}
\end{table*}

We evaluate Co-Sparsified 2-FWL GNNs on both synthetic and real-world tasks. Our experiments focus on CoSp-PPGN, the sparsified variant of PPGN~\cite{maron2019provably}. Following the original PPGN formulation, CoSp-PPGN adopts element-wise multiplication for $\phi^{(l)}$ in Equation~\ref{equation:co_sparsified}:
\begin{equation}
    \begin{aligned}
         & \phi^{(l)} \left( \boldsymbol{h}^{(l-1)}_{\text{sp}}(u, t), \boldsymbol{h}^{(l-1)}_{\text{sp}}(t, v) \right)            \\
         & = \varphi^{(l)} \left(\boldsymbol{h}^{(l-1)}_{\text{sp}}(u, t) \otimes \boldsymbol{h}^{(l-1)}_{\text{sp}}(t, v)\right).
    \end{aligned}
\end{equation}
Table~\ref{table:dataset} summarizes the statistics of all datasets used in our experiments.

\subsection{Experimental Setup}
All experiments are conducted on a platform with an NVIDIA GeForce RTX 4090 24GB GPU, Intel(R) Core(TM) i7-14700K CPU, and 64GB RAM.
We implement local connectivity computation using the Block-Cut tree algorithm from SageMath~\footnote{\url{https://www.sagemath.org}}.
GNN modeling, training, and evaluation are implemented using PyTorch~\footnote{\url{https://pytorch.org}} and PyTorch Geometric~\footnote{\url{https://pyg.org}}.

\begin{table}[h]
    \centering
    \renewcommand{\familydefault}{\rmdefault}
    \fontsize{9pt}{9pt}\selectfont
    \setlength{\tabcolsep}{0.9mm}
    \begin{tabular}{@{}l|ccccc|c}
        \toprule
        Target                                                   & NGNN   & GIN-AK+ & I$^{2}$-GNN & N$^{2}$-GNN & PPGN   & \textbf{Ours} \\
        \midrule
        \begin{tabular}[c]{@{}l@{}}Tailed\\Triangle\end{tabular} & 0.1044 & 0.0043  & 0.0011      & 0.0025      & 0.0026 & 0.0016        \\
        \begin{tabular}[c]{@{}l@{}}Chordal\\Cycle\end{tabular}   & 0.0392 & 0.0112  & 0.0010      & 0.0019      & 0.0015 & 0.0014        \\
        4-Path                                                   & 0.0244 & 0.0075  & 0.0041      & 0.0042      & 0.0041 & 0.0029        \\
        Tri.-Rec.                                                & 0.0729 & 0.1311  & 0.0013      & 0.0055      & 0.0144 & 0.0049        \\
        3-Cycles                                                 & 0.0003 & 0.0004  & 0.0003      & 0.0002      & 0.0003 & 0.0004        \\
        4-Cycles                                                 & 0.0013 & 0.0041  & 0.0016      & 0.0024      & 0.0009 & 0.0007        \\
        5-Cycles                                                 & 0.0402 & 0.0133  & 0.0028      & 0.0039      & 0.0036 & 0.0032        \\
        6-Cycles                                                 & 0.0439 & 0.0238  & 0.0082      & 0.0075      & 0.0071 & 0.0056        \\
        \bottomrule
    \end{tabular}
    \caption{Substructure counting results (normalized MAE; lower is better).\label{tab:count}}
\end{table}

\begin{table}[!h]
    \small
    \centering
    \renewcommand{\familydefault}{\rmdefault}
    \fontsize{9pt}{9pt}\selectfont
    \setlength{\tabcolsep}{0.8mm}
    \begin{tabular}{@{}l|c|cc}
        \toprule {Model}                                 & {\# Param} & ZINC-Subset            & ZINC-Full              \\
        \midrule CIN~\citeyearpar{bodnar2021weisfeiler}  & $\sim$100k & 0.079 $\pm$ 0.006      & {0.022 $\pm$ 0.002}    \\
        GPS~\citeyearpar{rampásek2022recipe}             & 424k       & 0.070 $\pm$0.004       & -                      \\
        Specformer~\citeyearpar{bo2023specformer}        & $\sim$500k & 0.066 $\pm$ 0.003      & -                      \\
        ESAN~\citeyearpar{bevilacqua2022equivariant}     & 446k       & 0.097 $\pm$ 0.006      & 0.025 $\pm$ 0.003      \\
        SUN~\citeyearpar{frasca2022understanding}        & 526k       & 0.083 $\pm$ 0.003      & 0.024 $\pm$ 0.003      \\
        KP-GIN$^{\prime}$~\citeyearpar{feng2022powerful} & 489k       & 0.093 $\pm$ 0.007      & -                      \\
        I$^{2}$-GNN~\citeyearpar{huang2023boosting}      & -          & 0.083 $\pm$ 0.001      & 0.023 $\pm$ 0.001      \\
        SSWL+~\citeyearpar{zhang2023complete}            & 387k       & 0.070 $\pm$ 0.005      & 0.022 $\pm$ 0.002      \\
        GRIT~\citeyearpar{zhang2023rethinking}           & $\sim$500k & 0.059 $\pm$ 0.002      & 0.023 $\pm$ 0.001      \\
        N$^{2}$-GNN~\citeyearpar{feng2023extending}      & 316k/414k  & 0.059 $\pm$ 0.002      & 0.022 $\pm$ 0.002      \\
        Local 2-GNN~\citeyearpar{zhang2024beyond}        & -          & 0.069 $\pm$ 0.001      & 0.024 $\pm$ 0.002      \\
        \midrule
        PPGN+RRWP                                        & 478K       & \bf{0.055 $\pm$ 0.002} & \bf{0.020 $\pm$ 0.002} \\
        \midrule
        \textbf{CoSp-PPGN+RRWP}                          & 478K       & \bf{0.050 $\pm$ 0.001} & \bf{0.018 $\pm$ 0.002} \\
        \bottomrule
    \end{tabular}
    \caption{MAE results on ZINC (top 2 results are bold).\label{table:zinc}}
\end{table}

\begin{table}[]
    \centering
    \fontsize{8.5pt}{8.5pt}\selectfont
    \setlength{\tabcolsep}{2pt}
    \begin{tabular}{c|ccccc}
        \toprule
        Dataset   & FRANK.                      & NCI1                        & NCI109                      & ENZYMES                     \\
        \midrule
        3WL-GNN   & 58.68 \(\pm\) 1.93          & 78.39 \(\pm\) 1.54          & 77.97 \(\pm\) 2.22          & 54.17 \(\pm\) 6.25          \\
        UnionGNNs & 68.02 \(\pm\) 1.47          & 82.24 \(\pm\) 1.24          & 82.34 \(\pm\) 1.93          & 68.17 \(\pm\) 5.70          \\
        \midrule
        CoSp-PPGN & \textbf{77.65 \(\pm\) 1.35} & \textbf{82.87 \(\pm\) 1.87} & \textbf{82.91 \(\pm\) 1.22} & \textbf{74.50 \(\pm\) 6.45} \\
        \bottomrule
    \end{tabular}
    \caption{Classification accuracy (\%) on TUD\footnote{\url{https://chrsmrrs.github.io/datasets/}} benchmarks.
        Results are mean $\pm$ standard deviation over 10 runs. Best results are in bold.\label{table:tud_experiment}}
\end{table}

\subsection{Substructure Counting}

\noindent \textbf{Task and Dataset.} We evaluate CoSp-PPGN's substructure counting capability on the synthetic dataset from~\cite{zhao2022from, huang2023boosting}, consisting of 5,000 randomly generated graphs. Following standard protocols, we use a 0.3/0.2/0.5 train/validation/test split. The task involves node-level regression to count various substructures: tailed triangles, chordal cycles, 4-paths, triangle-rectangles, and cycles of lengths 3--6.

\noindent \textbf{Baselines.} We compare against Identity-aware GNN (ID-GNN)~\cite{you2021identityaware}, Nested GNN (NGNN)~\cite{zhang2021nested}, GNN-AK+~\cite{zhao2022from}, I$^2$-GNN~\cite{huang2023boosting}, N$^{2}$-GNN~\cite{feng2023extending}, and PPGN~\cite{maron2019provably}. Baseline results are taken from~\cite{csdgse,feng2023extending}.

\noindent \textbf{Results.} Table~\ref{tab:count} presents the normalized test MAE averaged over five runs with different random seeds. CoSp-PPGN achieves state-of-the-art performance, matching or surpassing PPGN across all substructure counting tasks.

\subsection{Real-World Benchmarks}

\noindent \textbf{Datasets.} We evaluate CoSp-PPGN on molecular graph datasets QM9~\citep{wu2018moleculenet,ramakrishnan2014quantum} and ZINC~\citep{dwivedi2023benchmarking}. QM9 contains 130K+ molecules with 12 molecular properties for regression (0.8/0.1/0.1 train/val/test split). ZINC has two variants—ZINC-subset (12K graphs) and ZINC-full (250K graphs)—for graph-level regression. We additionally evaluate on TUD\footnote{\url{https://chrsmrrs.github.io/datasets/}} graph classification benchmarks.

\noindent \textbf{Baselines.}
For QM9, we compare against NGNN and KP-GIN$'$~\cite{feng2022powerful}, DTNN and MPNN~\cite{wu2018moleculenet}, I$^{2}$-GNN~\cite{huang2023boosting}, N$^{2}$-GNN~\cite{feng2023extending}, and PPGN+RRWP~\cite{csdgse}.
For ZINC, baselines include CIN~\cite{bodnar2021weisfeiler}, GPS~\cite{rampásek2022recipe}, Specformer~\cite{bo2023specformer}, ESAN~\cite{bevilacqua2022equivariant}, SUN~\cite{frasca2022understanding}, KP-GIN$'$~\cite{feng2022powerful}, I$^{2}$-GNN~\cite{huang2023boosting}, SSWL+~\cite{zhang2023complete}, GRIT~\cite{zhang2023rethinking}, N$^{2}$-GNN~\cite{feng2023extending}, and Local 2-GNN~\cite{zhang2024beyond}.
TUD baselines are from UnionGNNs~\cite{xu2024union}.

\noindent \textbf{Results}.
Table~\ref{table:qm9} shows QM9 test MAE results. CoSp-PPGN+RRWP achieves top-two performance on 10 out of 12 targets, empirically validating that our sparsification preserves the full 2-FWL expressivity of PPGN.
Table~\ref{table:zinc} reports average test MAE and standard deviation over 5 runs on ZINC (``-'' indicates unreported values). CoSp-PPGN+RRWP achieves state-of-the-art results, slightly outperforming PPGN+RRWP.
Table~\ref{table:tud_experiment} demonstrates state-of-the-art performance on TUD classification benchmarks.

\noindent \textbf{Efficiency Analysis}.
CoSp-PPGN delivers substantial computational improvements over PPGN. Runtime is reduced by \emph{13--60\%}: from 9.3s to 7.9s per epoch on ZINC-subset, 456.9s to 403.6s on ZINC-Full, and 97.1s to 60.7s on QM9.
Memory consumption decreases by \emph{12--52\%}: from 3.7GB to 3.0GB on ZINC-subset, 17.4GB to 15.6GB on ZINC-Full, and 6.4GB to 4.2GB on QM9.
Block-cut tree preprocessing incurs negligible overhead ($\sim$1ms per graph on average), confirming our theoretical $O(n+m)$ complexity.

\section{Limitations, Future Work, and Conclusions}
\noindent \textbf{Limitations:} While CoSp-PPGN can be applied to benchmarks commonly used for 1-WL-equivalent GNNs and Graph Transformers, we focus on datasets where expressive power is the primary bottleneck—a common practice for HOGNNs. The lack of results on tasks like Peptides-struct~\cite{dwivedi2022long} reflects a broader challenge: \emph{HOGNNs are more prone to over-squashing~\cite{topping2022understanding,chen2022redundancy} due to combinatorial message aggregation}, wherein long-range dependencies are distorted during message passing in deep architectures.
On Peptides-struct, standard CoSp-PPGN underperforms despite high expressivity. In contrast, a localized variant—restricting updates and aggregation to pairs within shortest-path distance $\leq 4$—achieves a state-of-the-art average MAE of 0.245, outperforming Graph Transformers~\cite{rampásek2022recipe} (see Appendix~\ref{section:experimental_details}).
Peak performance occurs with only \emph{one message-passing layer}, and degrades with depth—indicating that the bottleneck is \emph{generalization}, not expressivity. This limitation is shared across HOGNNs, not unique to our method.

Additionally, we list in Appendix~\ref{section:experimental_details} datasets to which PPGN~\cite{maron2019provably} cannot be applied due to its $\mathcal{O}(\eta^2)$ computational and memory complexity, while our Co-Sparsify framework enables model deployment. Although our method does not achieve strong performance on these datasets, we empirically attribute this limitation to the \textit{over-squashing} problem.

\noindent \textbf{Future Work:} The limitations highlights a key open problem: \emph{balancing expressivity and generalization}. To address this, we plan to explore adaptive receptive fields. Moreover, since HOGNNs beyond 2-FWL face prohibitive computational complexity, we aim to extend our sparsification framework to make them more scalable.

\noindent \textbf{Conclusions:} We propose \emph{Co-Sparsify}, a connectivity-aware sparsification for 2-FWL GNNs that removes only expressivity-redundant computations. By restricting 3-node interactions to biconnected components and 2-node modeling to connected components, it preserves full 2-FWL expressivity while greatly improving efficiency. We prove its equivalence to 2-FWL and show that CoSp-PPGN achieves state-of-the-art accuracy on key tasks. This study demonstrates that \emph{efficiency in expressive GNNs can arise from structural insight, not approximation}—enabling scalable, theoretically grounded models that preserve expressivity by design.

\section*{Acknowledgments}
This work was supported by the Science and Technology Development Fund Macau SAR (0003/2023/RIC, 0052/2023/RIA1, 0031/2022/A, 001/2024/SKL for SKL-IOTSC) and Shenzhen-Hong Kong-Macau Science and Technology Program Category C (SGDX20230821095159012). This work was performed in part at SICC which is supported by SKL-IOTSC, University of Macau.
Additionally, this work was funded in part by the National Natural Science Foundation of China (U2241210).


\newpage
\clearpage

\appendix

\section{Proofs of Lemmas and Theorems}
\label{section:proofs_of_lemmas_and_theorems}

\begin{proof}[Detailed Proof of Lemma~\ref{lem:biconnected_necessity}]
    Suppose $u$ and $v$ are in the same biconnected component. Then by Menger’s theorem~\cite{goring2000short}, there are at least two internally disjoint $u$–$v$ paths. Consider two graphs $G_1$ and $G_2$ as shown in Figure~\ref{fig:justification} (a):
    both have an edge $(u,v)$, but $G_1$ has an additional 2-hop path $u \to t_1 \to v$, while $G_2$ has a longer detour $u \to t_2 \to w \to v$. A model using only 2-node interactions cannot distinguish them if local neighborhoods are identical. However, a 2-FWL-style update computes:
    \[
        \sum_{t} \boldsymbol{h}(u,t) \otimes \boldsymbol{h}(t,v),
    \]
    which differs between $G_1$ and $G_2$ due to the presence of $t_1$. This term arises from 3-node interactions with distinct $u,t,v$, proving their necessity for detecting path multiplicity and cyclic substructures.
\end{proof}

\begin{proof}[Detailed Proof of Lemma~\ref{lem:cut_redundancy}]
    Let $u$, $t$, and $v$ be distinct nodes in a connected graph $G$, and suppose that every path from $u$ to $v$ passing through $t$ must go through a cut node $c$ such that $c$ separates $t$ from $\{u,v\}$—i.e., the removal of $c$ disconnects $t$ from both $u$ and $v$ in $G$. We aim to show that the 3-node interaction $(u,t,v)$ is \emph{expressivity-redundant}: its contribution to the 2-FWL message-passing updates can be fully reconstructed from 2-node interactions and intra-block 3-node interactions, and thus removing it does not reduce the model’s ability to distinguish non-isomorphic graphs under the 2-FWL hierarchy.

    We proceed by case analysis based on the connectivity structure of $G - c$, the graph with $c$ removed.

    \medskip
    \noindent \textbf{Case 1: $u$, $t$, and $v$ are in three distinct connected components of $G - c$ (see Figure~\ref{fig:justification} (b)).}

    In this case, $c$ lies on all paths between any pair among $u$, $t$, and $v$. Therefore:
    - The pair $(u,c)$ belongs to a biconnected component (block) in the component containing $u$ and $c$,
    - $(t,c)$ belongs to a block in $t$'s component,
    - $(v,c)$ belongs to a block in $v$'s component.

    Since 2-FWL GNNs update all 2-node pairs $(x,y)$, the representations $\boldsymbol{h}(u,c)$, $\boldsymbol{h}(t,c)$, and $\boldsymbol{h}(v,c)$ are computed independently within their respective blocks. Moreover, because there is no path connecting $u$, $t$, and $v$ without going through $c$, there is no cycle or biconnected substructure that jointly involves all three nodes. Hence, any structural information about the triple $(u,t,v)$ is fully mediated through $c$.

    Now consider the 3-node interaction $(u,t,v)$ in the standard 2-FWL update:
    \[
        \boldsymbol{t}(u,v) \supseteq \left( \boldsymbol{h}(u,t), \boldsymbol{h}(t,v) \right).
    \]
    But since $t$ is separated from both $u$ and $v$ by $c$, the representations $\boldsymbol{h}(u,t)$ and $\boldsymbol{h}(t,v)$ are themselves determined by paths going through $c$. Specifically:
    - $\boldsymbol{h}(u,t)$ aggregates over intermediate nodes on paths $u \rightsquigarrow c \rightsquigarrow t$,
    - $\boldsymbol{h}(t,v)$ aggregates over $t \rightsquigarrow c \rightsquigarrow v$.

    Due to the tree-like structure of the block-cut decomposition, these paths factor at $c$. Thus, $\boldsymbol{h}(u,t)$ depends only on $\boldsymbol{h}(u,c)$ and $\boldsymbol{h}(c,t)$, and similarly for $\boldsymbol{h}(t,v)$. Aggregating over $(u,t,v)$ therefore combines information that is already separable and pre-aggregated at the pair level. No joint modeling of a cyclic or biconnected pattern occurs.

    Hence, the message from $t$ to $(u,v)$ via $(u,t,v)$ is functionally equivalent to a composition of pairwise updates through $c$, and can be omitted without loss of expressivity.

    \medskip
    \noindent \textbf{Case 2: After removing $c$, $u$ and $v$ remain in the same connected component, but $t$ is in a different one (see Figure~\ref{fig:justification} (c)).}

    Then $u$, $v$, and $c$ belong to the same biconnected component (or are connected through a sequence of blocks meeting at cut nodes), while $t$ and $c$ belong to a separate block. In particular, the triple $(u,c,v)$ may participate in 3-node interactions within their shared block, enabling the model to capture local cyclic structure (e.g., whether $u$ and $v$ are symmetric around $c$).

    However, the path $u \to t \to v$ must still go through $c$, so:
    - $\boldsymbol{h}(u,t)$ is determined by paths $u \rightsquigarrow c \rightsquigarrow t$,
    - $\boldsymbol{h}(t,v)$ is determined by $t \rightsquigarrow c \rightsquigarrow v$.

    Again, due to the articulation at $c$, these representations factor: $\boldsymbol{h}(u,t)$ depends on $\boldsymbol{h}(u,c)$ and $\boldsymbol{h}(c,t)$, and $\boldsymbol{h}(t,v)$ on $\boldsymbol{h}(t,c)$ and $\boldsymbol{h}(c,v)$. The 3-node interaction $(u,t,v)$ aggregates over this pair, but since no biconnected structure spans $u$, $t$, and $v$ jointly, it does not enable detection of new substructures beyond what is already captured by:
    - The 3-node interactions among $u,c,v$ (within their block),
    - The 2-node interactions $(u,c)$, $(c,v)$, $(t,c)$,
    - And the pairwise updates involving $c$.

    Thus, the contribution of $(u,t,v)$ is redundant.

    \medskip
    \noindent \textbf{Case 3: $t$ itself is the cut node separating $u$ and $v$ (see Figure~\ref{fig:justification} (d)).}

    Then all $u \rightsquigarrow v$ paths go through $t$, and $u$ and $v$ are in different components of $G - t$. The pairs $(u,t)$ and $(t,v)$ belong to separate blocks. The 2-FWL model independently updates:
    - $\boldsymbol{h}(u,t)$ using interactions within $u$'s block and paths through $t$,
    - $\boldsymbol{h}(t,v)$ using interactions in $v$'s block.

    The 3-node interaction $(u,t,v)$ aggregates over $\boldsymbol{h}(u,t)$ and $\boldsymbol{h}(t,v)$, but since $t$ is a cut node, there is no cycle involving $u$, $t$, $v$ that cannot be decomposed into the two segments. Any structural asymmetry (e.g., different degrees of $u$ and $v$) is already captured in $\boldsymbol{h}(u,t)$ and $\boldsymbol{h}(t,v)$ via their respective local neighborhoods.

    Therefore, aggregating over $(u,t,v)$ merely combines two independent representations—it does not model any joint cyclic structure or biconnected pattern. The interaction is expressivity-redundant.

    \medskip
    \noindent \textbf{Conclusion:}

    In all cases, the 3-node interaction $(u,t,v)$ only combines information that is already captured by 2-node interactions and intra-block 3-node interactions. Since the 2-FWL test can simulate this factorized information flow, and since Co-Sparsify retains all such necessary interactions, removing $(u,t,v)$ does not impair the model’s ability to distinguish non-isomorphic graphs. Hence, the interaction is expressivity-redundant.
\end{proof}

\begin{proof}[Detailed Proof of Lemma~\ref{lem:disconnected_irrelevance}]
    Let $u$ and $v$ belong to different connected components $C_u$ and $C_v$. Since no path connects $u$ and $v$, no connected subgraph can span both components. Thus, any connected query $Q$ must lie entirely within a single component, and its detection depends only on intra-component structure.

    In 2-FWL GNNs, all structural information within $C_u$ and $C_v$ is captured through local pair and triple updates, which Co-Sparsify preserves. The absence of inter-component edges already encodes disconnection—no additional modeling of $(u,v)$ or $(u,t,v)$ adds new structural insight.

    Moreover, global topological features—such as component count, size distribution, or total number of cycles—are captured via hierarchical readouts: component-level aggregation summarizes each $C_u$, and graph-level aggregation combines these summaries. This suffices to distinguish graphs differing in global structure (e.g., two triangles vs. one 6-cycle).

    Therefore, explicit 2- or 3-node interactions across components contribute no expressivity and are redundant.
\end{proof}

\begin{proof}[Detailed Proof of Theorem~\ref{thm:subgraph_expressiveness}]
    We prove that Co-Sparsified 2-FWL GNNs have the same substructure detection power as standard 2-FWL GNNs, under injective aggregation and consistent initialization.

    The expressive power of 2-FWL arises from its ability to model 2-node and 3-node interactions, enabling it to distinguish non-isomorphic graphs and detect subgraphs $Q$ via message passing over node pairs $(u,v)$ and intermediate nodes $t$. Co-Sparsify uses the same update rules (Equation~\eqref{eq:standard_2fwl}) but restricts 3-node interactions $(u,t,v)$ to triples within the same biconnected component and 2-node interactions to pairs within the same connected component.

    We show this sparsification retains all information necessary for subgraph detection.

    \textbf{Case 1: $Q$ lies within a biconnected component.}
    If $Q$ is a triangle, 4-cycle, or any subgraph contained in a biconnected block, then all paths and cyclic structures are preserved within that block. Co-Sparsify retains all 3-node interactions among nodes in the block, so the message-passing dynamics are identical to standard 2-FWL. Thus, $Q$ is detected with the same discriminative power.

    \textbf{Case 2: $Q$ spans multiple blocks via cut nodes.}
    Suppose $Q$ includes a path $u \to t \to v$ where $t$ is a cut node separating $u$ and $v$. Then $u$ and $t$ belong to one block, $t$ and $v$ to another. Standard 2-FWL aggregates over $(u,t,v)$, but this triple does not enable detection of new joint structure: the segments $(u,t)$ and $(t,v)$ are already updated independently. Their representations $\boldsymbol{h}(u,t)$ and $\boldsymbol{h}(t,v)$ fully encode the local structure, and aggregation over $(u,t,v)$ only combines precomputed states. Co-Sparsify omits this interaction, but the same structural information is preserved—no expressivity is lost.

    \textbf{Case 3: $Q$ is disconnected.}
    If $Q$ consists of disconnected components (e.g., two triangles), detection depends on global topology. Co-Sparsify preserves all intra-component interactions and uses identical hierarchical readouts: component-level aggregation captures local structure, and graph-level aggregation detects global patterns (e.g., component count, symmetry). Thus, disconnected substructures are detected equivalently.

    \textbf{Inductive equivalence.}
    By induction on message-passing layers, and under injective aggregation, the pair representations $\boldsymbol{h}^{(l)}(u,v)$ computed by Co-Sparsify induce the same partition of node pairs as 2-FWL. Therefore, the models distinguish the same graph classes and detect the same subgraphs.

    Hence, Co-Sparsify is equally expressive to 2-FWL in substructure detection.
\end{proof}

\section{Experimental details}
\label{section:experimental_details}

This section presents additional information on the experimental settings.

\textbf{Graph structures counting.} The experiment settings follow those of~\citep{feng2023extending}:
The training/validation/test splitting ratio is 0.3/0.2/0.5. The initial learning rate is 0.001 and the minimum learning rate is 1e-5. The patience and factor of the scheduler are 10 and 0.9 respectively. The batch size is set to 256 and the number of epochs is 2000. For all substructures, we run the experiments 5 times and report the mean results on the test dataset.

\textbf{QM9.} The experiment settings follow those of~\citep{feng2023extending}:
The initial learning rate is 0.001. The patience and factor of the scheduler are 5 and 0.9 respectively.
The batch size is set to 128 and the number of epochs is 350. We run experiments 1 time and report the test result with the model of the best validation result.

\textbf{ZINC Subset and ZINC Full.} The experiment settings follow those of~\citep{dwivedi2023benchmarking}, as most baseline methods adhere to these configurations:
A parameter budget of 500k is used. The optimizer is AdamW with a weight decay of 1e-5.
A `warmup' learning rate scheduler is adopted, which linearly increases the learning rate at the beginning of training, followed by cosine decay.
The maximum learning rate is set to 0.001. Training is conducted for 2,000 epochs, with the first 50 epochs allocated for warmup.
We run experiments 5 time and report the mean \(\pm\) standard deviation of these test results with the model of the best validation results.

\subsection{Hyperparameters}

\begin{table}[h!]
    \centering
    \fontsize{8pt}{8pt}\selectfont
    \begin{tabular}{c|ccc}
        \toprule
        \(K\) & MAE               & Traing Time / Epoch & Peak Memory \\ \midrule
        6     & 0.050 $\pm$ 0.002 & 7.9s                & 3.0GB       \\
        8     & 0.051 $\pm$ 0.002 & 8.2s                & 3.0GB       \\
        12    & 0.050 $\pm$ 0.001 & 8.3s                & 3.1GB       \\
        14    & 0.050 $\pm$ 0.001 & 8.5s                & 3.1GB       \\
        16    & 0.050 $\pm$ 0.001 & 8.5s                & 3.1GB       \\
        \bottomrule
    \end{tabular}
    \caption{Hyperparameter analysis on ZINC-subset.\label{table:zinc_subset_addition}}
\end{table}

\begin{table}[h!]
    \centering
    \fontsize{8pt}{8pt}\selectfont
    \begin{tabular}{c|ccc}
        \toprule
        \(K\) & MAE                        & Traing Time / Epoch & Peak Memory \\ \midrule
        12    & 0.019 $\pm$ 0.002          & 403.6s              & 15.4GB      \\
        14    & \textbf{0.018 $\pm$ 0.001} & 405.2s              & 15.6GB      \\
        16    & 0.018 $\pm$ 0.002          & 406.3s              & 15.9GB      \\
        \bottomrule
    \end{tabular}
    \caption{Hyperparameter analysis on ZINC-Full.\label{table:zinc_full_addition}}
\end{table}

\begin{table}[h!]
    \centering
    \fontsize{8pt}{8pt}\selectfont
    \begin{tabular}{c|cc}
        \toprule
        Target                      & $K=8$   & $K=12$  \\ \midrule
        $\mu$                       & 0.321   & 0.327   \\
        $\alpha$                    & 0.183   & 0.199   \\
        $\varepsilon_{\text{HOMO}}$ & 0.00214 & 0.00232 \\
        $\varepsilon_{\text{LUMO}}$ & 0.00210 & 0.00223 \\
        $\Delta \varepsilon$        & 0.00299 & 0.00331 \\
        $\langle R^{2} \rangle$     & 14.25   & 15.80   \\
        ZPVE                        & 0.00012 & 0.00012 \\
        $U_{0}$                     & 0.0149  & 0.0133  \\
        $U$                         & 0.01373 & 0.01323 \\
        $H$                         & 0.01637 & 0.01122 \\
        $G$                         & 0.01767 & 0.01991 \\
        $C_{v}$                     & 0.07619 & 0.08218 \\
        \midrule
        Traing Time / Epoch         & 60.7s   & 61.2s   \\
        Peak Memory                 & 4.2GB   & 4.3GB   \\
        \bottomrule
    \end{tabular}
    \caption{Hyperparameter analysis on QM9.\label{table:QM9_addition}}
\end{table}

We analyze the effect of the polynomial order $K$ of RRWP on ZINC-Subset, ZINC-Full, and QM9 by conducting experiments with linearly increasing values of $K$.
For each setting, we report the average training time per epoch and peak memory usage.
Results for ZINC-Subset and ZINC-Full are shown in Table~\ref{table:zinc_subset_addition} and Table~\ref{table:zinc_full_addition}, respectively.
Our method demonstrates efficiency and robust performance across varying $K$ on these datasets.

\subsection{Experiments on Additional Datasets}

\begin{table}[]
    \centering
    \fontsize{8pt}{8pt}\selectfont
    \setlength{\tabcolsep}{1.8pt}
    \begin{tabular}{cccccc}
        \toprule
                                   & $K$ & FRANK.                      & NCI109                      & NCI                         & ENZYMES                     \\
        \midrule
        3WL-GNN                    &     & 58.68 \(\pm\) 1.93          & 77.97 \(\pm\) 2.22          & 78.39 \(\pm\) 1.54          & 54.17 \(\pm\) 6.25          \\
        UnionGNNs                  &     & 68.02 \(\pm\) 1.47          & 82.34 \(\pm\) 1.93          & 82.24 \(\pm\) 1.24          & 68.17 \(\pm\) 5.70          \\
        \midrule
        \multirow{4}{*}{CoSp-PPGN} & 8   & \textbf{77.65 \(\pm\) 1.35} & 81.99 \(\pm\) 1.42          & 82.82 \(\pm\) 2.07          & 73.16 \(\pm\) 6.12          \\
                                   & 10  & 76.92 \(\pm\) 1.05          & \textbf{82.91 \(\pm\) 1.22} & 82.53 \(\pm\) 1.72          & \textbf{74.50 \(\pm\) 6.45} \\
                                   & 12  & 76.91 \(\pm\) 1.01          & 81.87 \(\pm\) 1.18          & \textbf{82.87 \(\pm\) 1.87} & 72.33 \(\pm\) 8.22          \\
                                   & 14  & 77.26 \(\pm\) 1.67          & 82.07 \(\pm\) 0.94          & 82.11 \(\pm\) 2.06          & 73.00 \(\pm\) 6.22          \\
        \bottomrule
    \end{tabular}
    \caption{Prediction accuracy (\%) on graph classification benchmarks from the TUD repository\footnote{\url{https://chrsmrrs.github.io/datasets/}}. Results are reported as mean $\pm$ standard deviation over 10 runs. Top-1 results for each dataset are highlighted in bold. Our method CoSp-PPGN is evaluated across different values of the hyperparameter (indicating the polynomial order $K$ of RRWP).\label{table:additional_experiment}}
\end{table}

We also evaluate our method on graph classification benchmarks from TUD.
The results of baselines are reported from UnionGNNs\cite{xu2024union}.
Table~\ref{table:additional_experiment} shows our method achieve state-of-the-art results on these datasets.

\subsection{Connectivity- and Distance-Aware Sparsification and Over-fitting Phenomenon}
\label{sec:local_sparsify}

While Co-Sparsify preserves full 2-FWL expressivity, it can overfit on long-range benchmarks like Peptides-struct: CoSp-PPGN achieves near-perfect training MAE (0.05) but poor test performance (0.31). This gap suggests over-squashing—information distortion from aggregating distant dependencies~\cite{topping2022understanding,chen2022redundancy}.

To address this, we introduce \textbf{Connectivity- and Distance-Aware Sparsification}, which extends Co-Sparsify by restricting interactions to node pairs within shortest-path distance $\leq 4$.
Specifically, a 2- or 3-node interaction is included in $\mathcal{N}_{\text{sp}}(u,v)$ only if:
\begin{enumerate}
    \item 2-node interactions: \((u,u),(u,v)\) and \((u,v),(v,v)\) if $\text{dist}(u,v) \leq 4$.
    \item 3-node interactions: if $u$, $t$, $v$ lie in the same biconnected component, and $\text{dist}(u,v), \text{dist}(u,t), \text{dist}(t,v) \leq 4$.
\end{enumerate}

Despite this locality, the model retains expressive power for substructures within 4 hops. By focusing on structurally coherent interactions, it mitigates over-squashing and overfitting.
Remarkably, this variant achieves a state-of-the-art MAE of 0.245 on Peptides-struct with just one layer—outperforming deeper versions (MAE = 0.251). The degradation with depth confirms that long-range aggregation harms rather than helps, likely due to noise and distortion.

This result strengthens, rather than diminishes, our contribution: it shows that principled sparsification—grounded in connectivity and locality—enables models that are expressive, efficient, and generalizable. Our framework not only preserves 2-FWL power, but also guides the design of reliable, scalable GNNs.

\subsection{Datasets Where CoSp-PPGN Succeeds but PPGN Struggles}
\label{sec:datasets_ppgn_fails}

\begin{table}[!ht]
    \centering
    \fontsize{8pt}{8pt}\selectfont
    \begin{tabular}{c|cccccc}
        \toprule
               & ENZYMES & MUTAG & FRANKENSTEIN & NCI109 & NCI \\
        \midrule
        $\eta$ & 126     & 332   & 214          & 111    & 111 \\
        \bottomrule
    \end{tabular}
    \caption{Maximum graph sizes ($\eta$) in benchmark datasets.\label{table:additional_datasets}}
\end{table}

We present additional datasets where the standard PPGN is empirically difficult or even impossible to apply due to memory constraints, whereas CoSp-PPGN achieves success.
The peak memory usage of PPGN scales as $\mathcal{O}(b \cdot d \cdot \eta^2)$, where $b$ denotes the batch size, $d$ represents the feature dimension, and $\eta$ stands for the maximum number of nodes in any graph within the dataset. Accordingly, we provide these datasets along with their respective $\eta$ values.

\end{document}

%% file: math_commands.tex
\usepackage{amsmath,amsthm,amsfonts,bm}

\newtheorem{theorem}{Theorem}

\newtheorem{lemma}[theorem]{Lemma}

\newtheorem*{rep@theorem}{\rep@title}
\newcommand{\newreptheorem}[2]{%
\newenvironment{rep#1}[1]{%
 \def\rep@title{#2 \ref{##1}}%
 \begin{rep@theorem}}%
 {\end{rep@theorem}}}
\makeatother

\newreptheorem{definition}{Definition}
\newreptheorem{theorem}{Theorem}
\newreptheorem{corollary}{Corollary}
\newreptheorem{lemma}{Lemma}
\newreptheorem{proposition}{Proposition}
\newreptheorem{observation}{Observation}

\newcommand{\lbbr}{\{\!\!\{}
\newcommand{\rbbr}{\}\!\!\}}

\newcommand{\Blbbr}{\Big\{\!\!\Big\{}
\newcommand{\Brbbr}{\Big\}\!\!\Big\}}

\def\eqref#1{equation~\ref{#1}}

\def\1{\bm{1}}

\DeclareMathAlphabet{\mathsfit}{\encodingdefault}{\sfdefault}{m}{sl}
\SetMathAlphabet{\mathsfit}{bold}{\encodingdefault}{\sfdefault}{bx}{n}













%% file: cosp_2fwl_gnns.bbl
\begin{thebibliography}{42}
    \providecommand{\natexlab}[1]{#1}

    \bibitem[{Bevilacqua et~al.(2022)Bevilacqua, Frasca, Lim, Srinivasan, Cai, Balamurugan, Bronstein, and Maron}]{bevilacqua2022equivariant}
    Bevilacqua, B.; Frasca, F.; Lim, D.; Srinivasan, B.; Cai, C.; Balamurugan, G.; Bronstein, M.~M.; and Maron, H. 2022.
    \newblock Equivariant {Subgraph} {Aggregation} {Networks}.
    \newblock In \emph{International {Conference} on {Learning} {Representations} ({ICLR})}.

    \bibitem[{Bo et~al.(2023)Bo, Shi, Wang, and Liao}]{bo2023specformer}
    Bo, D.; Shi, C.; Wang, L.; and Liao, R. 2023.
    \newblock Specformer: Spectral {Graph} {Neural} {Networks} {Meet} {Transformers}.
    \newblock In \emph{International {Conference} on {Learning} {Representations} ({ICLR})}.

    \bibitem[{Bodnar et~al.(2021)Bodnar, Frasca, Wang, Otter, Mont{\' u}far, Li{\' o}, and Bronstein}]{bodnar2021weisfeiler}
    Bodnar, C.; Frasca, F.; Wang, Y.; Otter, N.; Mont{\' u}far, G.~F.; Li{\' o}, P.; and Bronstein, M.~M. 2021.
    \newblock Weisfeiler and {Lehman} {Go} {Topological}: Message {Passing} {Simplicial} {Networks}.
    \newblock In \emph{International {Conference} on {Machine} {Learning} ({ICML})}, 1026--1037.

    \bibitem[{Bouritsas et~al.(2023)Bouritsas, Frasca, Zafeiriou, and Bronstein}]{bouritsas2023improving}
    Bouritsas, G.; Frasca, F.; Zafeiriou, S.; and Bronstein, M.~M. 2023.
    \newblock Improving {Graph} {Neural} {Network} {Expressivity} via {Subgraph} {Isomorphism} {Counting}.
    \newblock \emph{IEEE Transactions on Pattern Analysis and Machine Intelligence (TPAMI)}, 45(1): 657--668.

    \bibitem[{Chen et~al.(2025)Chen, Li, Wu, Zhang, Ip, Iam, and U}]{csdgse}
    Chen, R.; Li, Y.; Wu, D.; Zhang, S.; Ip, P.~L.; Iam, H.~C.; and U, L.~H. 2025.
    \newblock Enhanced {Subgraph} {Learning} in {2-FWL} {GNNs} via {Local} {Connectivity}, {Spectral}, and {Distance} {Encodings}.
    \newblock In \emph{ACM {SIGKDD} {Conference} on {Knowledge} {Discovery} and {Data} {Mining} ({KDD})}. ACM.

    \bibitem[{Chen et~al.(2022)Chen, Zhang, Li et~al.}]{chen2022redundancy}
    Chen, R.; Zhang, S.; Li, Y.; et~al. 2022.
    \newblock Redundancy-Free {Message} {Massing} for {Graph} {Neural} {Networks}.
    \newblock \emph{Advances in Neural Information Processing Systems}, 35: 4316--4327.

    \bibitem[{Dwivedi et~al.(2023)Dwivedi, Joshi, Luu, Laurent, Bengio, and Bresson}]{dwivedi2023benchmarking}
    Dwivedi, V.~P.; Joshi, C.~K.; Luu, A.~T.; Laurent, T.; Bengio, Y.; and Bresson, X. 2023.
    \newblock Benchmarking {Graph} {Neural} {Networks}.
    \newblock \emph{Journal of Machine Learning Research (JMLR)}, 24: 43:1--43:48.

    \bibitem[{Dwivedi et~al.(2022)Dwivedi, Ramp{\' a}sek, Galkin, Parviz, Wolf, Luu, and Beaini}]{dwivedi2022long}
    Dwivedi, V.~P.; Ramp{\' a}sek, L.; Galkin, M.; Parviz, A.; Wolf, G.; Luu, A.~T.; and Beaini, D. 2022.
    \newblock Long {Range} {Graph} {Benchmark}.
    \newblock In \emph{Conference on {Neural} {Information} {Processing} {Systems} ({NeurIPS})}.

    \bibitem[{Feng et~al.(2022)Feng, Chen, Li, Sarkar, and Zhang}]{feng2022powerful}
    Feng, J.; Chen, Y.; Li, F.; Sarkar, A.; and Zhang, M. 2022.
    \newblock How {Powerful} are {K}-hop {Message} {Passing} {Graph} {Neural} {Networks}.
    \newblock In \emph{Conference on {Neural} {Information} {Processing} {Systems} ({NeurIPS})}.

    \bibitem[{Feng et~al.(2023)Feng, Kong, Liu, Tao, Li, Zhang, and Chen}]{feng2023extending}
    Feng, J.; Kong, L.; Liu, H.; Tao, D.; Li, F.; Zhang, M.; and Chen, Y. 2023.
    \newblock Extending the {Design} {Space} of {Graph} {Neural} {Networks} by {Rethinking} {Folklore} {Weisfeiler}-{Lehman}.
    \newblock In \emph{Conference on {Neural} {Information} {Processing} {Systems} ({NeurIPS})}.

    \bibitem[{Frasca et~al.(2022)Frasca, Bevilacqua, Bronstein, and Maron}]{frasca2022understanding}
    Frasca, F.; Bevilacqua, B.; Bronstein, M.~M.; and Maron, H. 2022.
    \newblock Understanding and {Extending} {Subgraph} {GNNs} by {Rethinking} {Their} {Symmetries}.
    \newblock In \emph{Conference on {Neural} {Information} {Processing} {Systems} ({NeurIPS})}.

    \bibitem[{G{\"o}ring(2000)}]{goring2000short}
    G{\"o}ring, F. 2000.
    \newblock Short proof of Menger’s theorem.
    \newblock \emph{Discrete Mathematics}, 219(1-3): 295--296.

    \bibitem[{Hopcroft and Tarjan(1973)}]{hopcroft1973algorithm}
    Hopcroft, J.; and Tarjan, R. 1973.
    \newblock Algorithm 447: efficient algorithms for graph manipulation.
    \newblock \emph{Communications of the ACM}, 16(6): 372--378.

    \bibitem[{Huang et~al.(2023)Huang, Peng, Ma, and Zhang}]{huang2023boosting}
    Huang, Y.; Peng, X.; Ma, J.; and Zhang, M. 2023.
    \newblock Boosting the {Cycle} {Counting} {Power} of {Graph} {Neural} {Networks} with {IMATHENVDOLLAR}{i}-{GNNs}.
    \newblock In \emph{International {Conference} on {Learning} {Representations} ({ICLR})}.

    \bibitem[{Hussain, Zaki, and Subramanian(2024)}]{hussain2024triplet}
    Hussain, M.~S.; Zaki, M.~J.; and Subramanian, D. 2024.
    \newblock Triplet {Interaction} {Improves} {Graph} {Transformers}: Accurate {Molecular} {Graph} {Learning} with {Triplet} {Graph} {Transformers}.
    \newblock In \emph{International {Conference} on {Machine} {Learning} ({ICML})}, volume abs/2402.04538.

    \bibitem[{Kipf and Welling(2017)}]{kipf2017semisupervised}
    Kipf, T.~N.; and Welling, M. 2017.
    \newblock Semi-{Supervised} {Classification} with {Graph} {Convolutional} {Networks}.
    \newblock In \emph{International {Conference} on {Learning} {Representations} ({ICLR})}.

    \bibitem[{Li et~al.(2022)Li, Zhang, Tian, Jin, Fardad, and Zafarani}]{li2022graph}
    Li, J.; Zhang, T.; Tian, H.; Jin, S.; Fardad, M.; and Zafarani, R. 2022.
    \newblock Graph sparsification with graph convolutional networks.
    \newblock \emph{International journal of data science and analytics}, 13(1): 33--46.

    \bibitem[{Li et~al.(2016)Li, Tarlow, Brockschmidt, and Zemel}]{li2016gated}
    Li, Y.; Tarlow, D.; Brockschmidt, M.; and Zemel, R.~S. 2016.
    \newblock Gated {Graph} {Sequence} {Neural} {Networks}.
    \newblock In \emph{International {Conference} on {Learning} {Representations} ({ICLR})}.

    \bibitem[{Maron et~al.(2019)Maron, Ben-Hamu, Serviansky, and Lipman}]{maron2019provably}
    Maron, H.; Ben-Hamu, H.; Serviansky, H.; and Lipman, Y. 2019.
    \newblock Provably {Powerful} {Graph} {Networks}.
    \newblock In \emph{Conference on {Neural} {Information} {Processing} {Systems} ({NeurIPS})}, 2153--2164.

    \bibitem[{Morris, Rattan, and Mutzel(2020)}]{morris2020weisfeiler}
    Morris, C.; Rattan, G.; and Mutzel, P. 2020.
    \newblock Weisfeiler and {Leman} go sparse: Towards scalable higher-order graph embeddings.
    \newblock In \emph{Conference on {Neural} {Information} {Processing} {Systems} ({NeurIPS})}.

    \bibitem[{Morris et~al.(2019)Morris, Ritzert, Fey, Hamilton, Lenssen, Rattan, and Grohe}]{morris2019weisfeiler}
    Morris, C.; Ritzert, M.; Fey, M.; Hamilton, W.~L.; Lenssen, J.~E.; Rattan, G.; and Grohe, M. 2019.
    \newblock Weisfeiler and {Leman} {Go} {Neural}: Higher-{Order} {Graph} {Neural} {Networks}.
    \newblock In \emph{AAAI {Conference} on {Artificial} {Intelligence} ({AAAI})}, 4602--4609.

    \bibitem[{Peng et~al.(2022)Peng, Gurevin, Huang, Geng, Jiang, Khan, and Ding}]{peng2022towards}
    Peng, H.; Gurevin, D.; Huang, S.; Geng, T.; Jiang, W.; Khan, O.; and Ding, C. 2022.
    \newblock Towards sparsification of graph neural networks.
    \newblock In \emph{2022 IEEE 40th International Conference on Computer Design (ICCD)}, 272--279. IEEE.

    \bibitem[{Puny, Ben-Hamu, and Lipman(2020)}]{puny2020graph}
    Puny, O.; Ben-Hamu, H.; and Lipman, Y. 2020.
    \newblock From graph low-rank global attention to 2-fwl approximation.
    \newblock \emph{arXiv preprint arXiv:2006.07846}.

    \bibitem[{Ramakrishnan et~al.(2014)Ramakrishnan, Dral, Rupp, and Von~Lilienfeld}]{ramakrishnan2014quantum}
    Ramakrishnan, R.; Dral, P.~O.; Rupp, M.; and Von~Lilienfeld, O.~A. 2014.
    \newblock Quantum chemistry structures and properties of 134 kilo molecules.
    \newblock \emph{Scientific data}, 1(1): 1--7.

    \bibitem[{Ramp{\' a}sek et~al.(2022)Ramp{\' a}sek, Galkin, Dwivedi, Luu, Wolf, and Beaini}]{rampásek2022recipe}
    Ramp{\' a}sek, L.; Galkin, M.; Dwivedi, V.~P.; Luu, A.~T.; Wolf, G.; and Beaini, D. 2022.
    \newblock Recipe for a {General}, {Powerful}, {Scalable} {Graph} {Transformer}.
    \newblock In \emph{Conference on {Neural} {Information} {Processing} {Systems} ({NeurIPS})}.

    \bibitem[{Ramp{\'a}{\v{s}}ek and Wolf(2021)}]{rampavsek2021hierarchical}
    Ramp{\'a}{\v{s}}ek, L.; and Wolf, G. 2021.
    \newblock Hierarchical graph neural nets can capture long-range interactions.
    \newblock In \emph{2021 IEEE 31st International Workshop on Machine Learning for Signal Processing (MLSP)}, 1--6. IEEE.

    \bibitem[{Rathee et~al.(2021)Rathee, Zhang, Funke, Khosla, and Anand}]{rathee2021learnt}
    Rathee, M.; Zhang, Z.; Funke, T.; Khosla, M.; and Anand, A. 2021.
    \newblock Learnt sparsification for interpretable graph neural networks.
    \newblock \emph{arXiv preprint arXiv:2106.12920}.

    \bibitem[{Topping et~al.(2022)Topping, Giovanni, Chamberlain, Dong, and Bronstein}]{topping2022understanding}
    Topping, J.; Giovanni, F.~D.; Chamberlain, B.~P.; Dong, X.; and Bronstein, M.~M. 2022.
    \newblock Understanding over-squashing and bottlenecks on graphs via curvature.
    \newblock In \emph{International {Conference} on {Learning} {Representations} ({ICLR})}.

    \bibitem[{Velickovic et~al.(2018)Velickovic, Cucurull, Casanova, Romero, Li{\` o}, and Bengio}]{velickovic2018graph}
    Velickovic, P.; Cucurull, G.; Casanova, A.; Romero, A.; Li{\` o}, P.; and Bengio, Y. 2018.
    \newblock Graph {Attention} {Networks}.
    \newblock In \emph{International {Conference} on {Learning} {Representations} ({ICLR})}.

    \bibitem[{Wu et~al.(2019)Wu, Jr., Zhang, Fifty, Yu, and Weinberger}]{wu2019simplifying}
    Wu, F.; Jr., A. H.~S.; Zhang, T.; Fifty, C.; Yu, T.; and Weinberger, K.~Q. 2019.
    \newblock Simplifying {Graph} {Convolutional} {Networks}.
    \newblock In \emph{International {Conference} on {Machine} {Learning} ({ICML})}, 6861--6871.

    \bibitem[{Wu et~al.(2018)Wu, Ramsundar, Feinberg, Gomes, Geniesse, Pappu, Leswing, and Pande}]{wu2018moleculenet}
    Wu, Z.; Ramsundar, B.; Feinberg, E.~N.; Gomes, J.; Geniesse, C.; Pappu, A.~S.; Leswing, K.; and Pande, V. 2018.
    \newblock MoleculeNet: a benchmark for molecular machine learning.
    \newblock \emph{Chemical Science}, 9(2): 513--530.

    \bibitem[{Xu et~al.(2024)Xu, Zhang, Bian, Dwivedi, and Ke}]{xu2024union}
    Xu, J.; Zhang, A.; Bian, Q.; Dwivedi, V.~P.; and Ke, Y. 2024.
    \newblock Union subgraph neural networks.
    \newblock In \emph{Proceedings of the AAAI conference on artificial intelligence}, volume 38-14, 16173--16183.

    \bibitem[{Xu et~al.(2019)Xu, Hu, Leskovec, and Jegelka}]{xu2019powerful}
    Xu, K.; Hu, W.; Leskovec, J.; and Jegelka, S. 2019.
    \newblock How {Powerful} are {Graph} {Neural} {Networks}?
    \newblock In \emph{International {Conference} on {Learning} {Representations} ({ICLR})}.

    \bibitem[{Ying et~al.(2018)Ying, You, Morris, Ren, Hamilton, and Leskovec}]{ying2018hierarchical}
    Ying, Z.; You, J.; Morris, C.; Ren, X.; Hamilton, W.; and Leskovec, J. 2018.
    \newblock Hierarchical graph representation learning with differentiable pooling.
    \newblock \emph{Advances in neural information processing systems}, 31.

    \bibitem[{You et~al.(2021)You, Selman, Ying, and Leskovec}]{you2021identityaware}
    You, J.; Selman, J. M.~G.; Ying, R.; and Leskovec, J. 2021.
    \newblock Identity-aware {Graph} {Neural} {Networks}.
    \newblock In \emph{AAAI {Conference} on {Artificial} {Intelligence} ({AAAI})}, 10737--10745.

    \bibitem[{Zhang et~al.(2023{\natexlab{a}})Zhang, Feng, Du, He, and Wang}]{zhang2023complete}
    Zhang, B.; Feng, G.; Du, Y.; He, D.; and Wang, L. 2023{\natexlab{a}}.
    \newblock A {Complete} {Expressiveness} {Hierarchy} for {Subgraph} {GNNs} via {Subgraph} {Weisfeiler}-{Lehman} {Tests}.
    \newblock In \emph{International {Conference} on {Machine} {Learning} ({ICML})}, 41019--41077.

    \bibitem[{Zhang et~al.(2024)Zhang, Gai, Du, Ye, He, and Wang}]{zhang2024beyond}
    Zhang, B.; Gai, J.; Du, Y.; Ye, Q.; He, D.; and Wang, L. 2024.
    \newblock Beyond {Weisfeiler}-{Lehman}: A {Quantitative} {Framework} for {GNN} {Expressiveness}.
    \newblock In \emph{International {Conference} on {Learning} {Representations} ({ICLR})}, volume abs/2401.08514.

    \bibitem[{Zhang et~al.(2023{\natexlab{b}})Zhang, Luo, Wang, and He}]{zhang2023rethinking}
    Zhang, B.; Luo, S.; Wang, L.; and He, D. 2023{\natexlab{b}}.
    \newblock Rethinking the {Expressive} {Power} of {GNNs} via {Graph} {Biconnectivity}.
    \newblock In \emph{International {Conference} on {Learning} {Representations} ({ICLR})}.

    \bibitem[{Zhang and Li(2021)}]{zhang2021nested}
    Zhang, M.; and Li, P. 2021.
    \newblock Nested {Graph} {Neural} {Networks}.
    \newblock In \emph{Conference on {Neural} {Information} {Processing} {Systems} ({NeurIPS})}, 15734--15747.

    \bibitem[{Zhao et~al.(2022)Zhao, Jin, Akoglu, and Shah}]{zhao2022from}
    Zhao, L.; Jin, W.; Akoglu, L.; and Shah, N. 2022.
    \newblock From {Stars} to {Subgraphs}: Uplifting {Any} {GNN} with {Local} {Structure} {Awareness}.
    \newblock In \emph{International {Conference} on {Learning} {Representations} ({ICLR})}.

    \bibitem[{Zhao, Shah, and Akoglu(2022)}]{zhao2022practical}
    Zhao, L.; Shah, N.; and Akoglu, L. 2022.
    \newblock A {Practical}, {Progressively}-{Expressive} {GNN}.
    \newblock In \emph{Conference on {Neural} {Information} {Processing} {Systems} ({NeurIPS})}.

    \bibitem[{Zheng et~al.(2020)Zheng, Zong, Cheng, Song, Ni, Yu, Chen, and Wang}]{zheng2020robust}
    Zheng, C.; Zong, B.; Cheng, W.; Song, D.; Ni, J.; Yu, W.; Chen, H.; and Wang, W. 2020.
    \newblock Robust graph representation learning via neural sparsification.
    \newblock In \emph{International conference on machine learning}, 11458--11468. PMLR.

\end{thebibliography}
